\documentclass[12pt]{amsart}
\usepackage{amsmath, amssymb, amsthm, amsfonts, enumerate, verbatim, mathtools,enumitem, bbm, bm}
\usepackage{pgf,tikz,pgfplots}
\usepackage[justification=centering]{caption}
\pgfplotsset{compat=1.15}
\usepackage{mathrsfs}
\usepackage{graphicx}
\usetikzlibrary{arrows,positioning}
\usepackage[all]{xy}
\usepackage{tikz}
\usepackage{subfig}
\usepackage{xcolor, soul, todonotes}
\usepackage[normalem]{ulem}
\usepackage{listings}
\usepackage{diagbox}
\usepackage[hypertexnames=false,debug,linktocpage=true,hidelinks]{hyperref}

\hypersetup{
	colorlinks,
	linktoc=all,
	linkcolor={blue},
	citecolor={red},
	urlcolor={blue}
}
\tikzstyle{punkt}=[circle, fill=black, minimum size=1mm,inner sep=0pt, draw]

\newtheorem{Theorem}{Theorem}[section]

\newtheorem{Corollary}[Theorem]{Corollary}

\newtheorem{Remark}[Theorem]{Remark}

\newtheorem{Example}[Theorem]{Example}

\newtheorem{Definition}[Theorem]{Definition}

\newtheorem{Question}[Theorem]{Question}

\let\epsilon\varepsilon
\let\kappa=\varkappa
\def\pnt{{\raise0.5mm\hbox{\large\bf.}}}

\newif\ifistoreview
\istoreviewtrue

\newcommand{\Do}{\mathrm{do}}

\newcommand{\kl}{\mathrm{KL}}
\renewcommand{\d}{\mathrm{d}}
\renewcommand{\P}{\mathbb{P}}
\newcommand{\E}{\mathbb{E}}

\newcommand{\NATE}{\mathrm{NATE}}

\newcommand{\FATE}{\mathrm{FATE}}
\newcommand{\GFATE}{\mathrm{GFATE}}
\newcommand{\NFATE}{\mathrm{NFATE}}
\newcommand{\NGFATE}{\mathrm{NGFATE}}
\newcommand{\ATE}{\mathrm{ATE}}
\begin{document}
	\title{Integrating Fuzzy Logic with Causal Inference: Enhancing the Pearl and Neyman-Rubin Methodologies}
\author[Amir Saki]{Amir Saki}
\address{Amir Saki, Department of Mathematics and Computer Science, The University of Quebec at Trois-Rivieres, 3351 Bd des Forges, Trois-Rivières, QC G8Z 4M3, Canada}
\email{amir.saki@uqtr.ca, amir.saki.math@gmail.com}
\author[Usef Faghihi]{Usef Faghihi\,*}
\thanks{* Corresponding author}
\address{Usef Faghihi, Department of Mathematics and Computer Science, The University of Quebec at Trois-Rivieres, 3351 Bd des Forges, Trois-Rivières, QC G8Z 4M3, Canada.}
\email{usef.faghihi@uqtr.ca}

	\begin{abstract}
		In this paper, we generalize the  Pearl and Neyman-Rubin  methodologies in causal inference by introducing a generalized approach that incorporates fuzzy logic. 
	Indeed, we inroduce a fuzzy causal inference approach that consider both the vagueness and imprecision inherent in data, as well as the subjective human perspective characterized by fuzzy terms such as 'high', 'medium', and 'low'. To do so, we introduce two fuzzy causal effect formulas: the Fuzzy Average Treatment Effect (FATE) and the Generalized Fuzzy Average Treatment Effect (GFATE), together with their normalized versions: NFATE and NGFATE.  When dealing with a binary treatment variable, our fuzzy causal effect formulas coincide with classical Average Treatment Effect (ATE) formula, that is a well-established and popular metric  in causal inference. In FATE, all values of the treatment variable are considered equally important. In contrast, GFATE  takes into account the rarity and frequency of these values.
	 We show that for linear Structural Equation Models (SEMs), the normalized versions of our formulas, NFATE and NGFATE, are equivalent to  ATE. Further, we provide identifiability criteria for these formulas and show their stability with respect to minor variations in the fuzzy subsets and the probability distributions involved. This ensures the robustness of our approach in handling small perturbations in the data. Finally, we provide several experimental examples to empirically validate and demonstrate the practical application of our proposed fuzzy causal inference methods.
	\end{abstract}

	
	\keywords{Causal inference, Pearl graphical model, structural equation models, fuzzy causal inference, fuzzy average treatment effect}

\maketitle	
\section{Introduction}

In the complex domain of causal inference, the challenge has always been to extract clear insights from the often vague and imprecise data coming from real-world scenarios. Notable approaches to causal inference are dominated by the frameworks proposed by Pearl and Neyman-Rubin. These frameworks have provided robust methodologies and points of views to find the cause-and-effect relationships from both observational and experimental data. However, these frameworks sometimes are in trouble for dealing with the inherent fuzziness of real-life data and the subjectivity of the human decision-making.

This paper introduces a new framework that extends the classical causal inference frameworks by integrating  fuzzy logic to handling data ambiguity and imprecision. Our approach includes introducing two new causal effect metrics—Fuzzy Average Treatment Effect (FATE) and Generalized Fuzzy Average Treatment Effect (GFATE)—along with their normalized counterparts. These metrics are designed to consider  binary treatment scenarios to more complex fuzzy environments.

 In FATE, all values of the treatment variable are assumed equally important. For instance, whether a treatment value is rarely or frequently observed, FATE considers both with the same importance. This is because intervention isolates the treatment variable from the influence of all other factors, which makes   it uniformly distributed. However, GFATE takes into account the rarity and frequency of these values, and hence it can adjust the treatment values accordingly during intervention.

By considering the causal inference problem through the lens of fuzzy logic, our aim is to fill the gap between quantitative precision and qualitative flexibility. This allows for an  interpretation of causal effects that respects the complexity of human language and thought processes, such as the use of terms like 'high', 'medium', and 'low'. Furthermore, our paper determines a rigorous mathematical foundation for the aforementioned fuzzy causal metrics, ensuring their stability and robustness with respect to identifiability criteria and sensitivity analysis.

Previous research on integrating fuzzy concepts with causal inference has often focused on causal discovery or  identifying the degrees of fuzzy causal relationships between variables by using various algorithms. However, in these studies,  a fuzzy causal effect metric is often not explicitly defined or developed  through a rigorous mathematical approach. 

Instead, the concentration has generally been on algorithmic exploration of the relationships, leaving a gap in the formal mathematical formulation and comprehensive theoretical development of such metrics. This gap indicates a potential area for further investigation and development in the field of fuzzy causal inference. 

For instance, in \cite{zhou2006fuzzy},  the authors introduce a  framework for modeling causal relationships with fuzzy logic, focusing on the evolution of causal influences within a network structure over time. This approach specifically  emphasizes on the network's ability to reach a convergence through adaptive learning of the relationships and strengths between various entities.
The paper \cite{miao2000causal}  illustrates the structural complexities and computational challenges of Fuzzy Cognitive Maps (FCMs)—dynamic systems where nodes and edges represent concepts and  the degrees of causal influence, respectively.
Also, \cite{schneider2016case} explores integrating fuzzy-set Qualitative Comparative Analysis (QCA) with case studies by emphasizing on the importance of carefully selecting cases based on specific criteria to ensure they meaningfully contribute to understanding causal mechanisms. 
The paper \cite{wang2014fuzzy} introduces a formal methodology to deal with ambiguity and imprecision in language through fuzzy algebra, which aides cognitive systems to mimic human-like reasoning and problem-solving processes. The authors of \cite{kunitomo2022causal} introduce a method to explore  causal relationships between fuzzified inputs and outputs by using algorithms such as Peter-Clark and Fast Causal Inference, combined with entropy-based testing to avoid data distribution assumptions.

The current research builds upon the conceptual work presented in \cite{faghihi2020association} and fundamental works in \cite{saki2022fundamental} and \cite{sakifaghihi}, where the authors for the first time explored the integration of fuzzy concepts with causal inference, aiming to address the inherent imprecision of real-world data and particularly focusing on developing formulas for measuring fuzzy-type causal effects.

The structure of this paper is organized as follows: Section \ref{Preliminaries} reviews the necessary preliminaries, including a brief overview of fuzzy logic and its applications to causal inference. Section \ref{Main Results} presents our main results, detailing the properties and applications of FATE. Section \ref{Generalized Fuzzy Average Treatment Effect} is devoted to our main results, detailing the properties and applications of FATE. In Section \ref{Causality in Fuzzy Systems}, we explore the integration of FATE and GFATE with fuzzy systems and probabilistic fuzzy systems, delving into how these metrics can be effectively applied within such frameworks. Finally, Section \ref{Examples} illustrates the practical utility of our approach through several experimental examples, demonstrating how fuzzy causal inference can be effectively applied in diverse scenarios.

By advancing this fuzzy logic-based framework, we not only enhance the toolkit available for causal analysis but also provide a means to make these tools more adaptable to the complexities and uncertainties of real-world data, thereby opening new avenues for research and application in the fields of statistics, economics, health sciences, and beyond.

\section{Preliminaries}\label{Preliminaries}
\subsection{Fuzzy Logics}
Such a logic is an approach to determine the ``degrees of truth" rather than the usual ``true or false" (1 or 0) binary logic. Fuzzy logic, as an extension to classical set theory, is particularly useful in dealing with imprecise or vague data, common in real-world scenarios.
In classical set theory, an element either belongs to a set or does not. Fuzzy subsets, however, introduce the concept of partial membership. This means that an element can belong to a set to a certain degree, ranging between 0 (completely outside the set) and 1 (completely inside the set).
For example, consider the set of "tall people." In a fuzzy subset, rather than classifying each person as either tall or not tall, people can belong to the set of tall people to varying degrees, such as 0.7 or 0.4, depending on their height.
The degree to which an element belongs to a fuzzy subset is quantified using a \textit{membership function}. This function maps each element to a membership value between 0 and 1. The shape of the membership function can vary depending on the context and the nature of the data. Common shapes for membership functions include triangular, trapezoidal, and bell-shaped (Gaussian) curves. By a \textit{fuzzy attribute} of a set $S \subseteq \mathbb{R}$ we refer to a fuzzy subset characterized by a membership function $f: S \rightarrow [0,1]$ in such a way that there exists $s\in S$ with $f(s)=0$. For insights into fuzzy logic and its applications, consult \cite{klir1995fuzzy, klir1996fuzzy, ross2009fuzzy, zadeh1965fuzzy, zadeh1974fuzzy}. 

\subsection{Fuzzy Rule-Based Systems}
They are a type of decision-making framework that uses fuzzy logic to interpret and then process data. These systems are designed to mimic human reasoning by handling imprecise or uncertain information, making them particularly effective in complex or poorly-defined problem spaces where traditional binary logic systems fall short.

The core of a fuzzy rule-based system is a set of fuzzy rules. These rules are formulated in an ``IF-THEN" structure, similar to human reasoning patterns. For example, a simple fuzzy rule might be: ``IF the temperature is high, THEN  ice cream sales should be high."These rules might come from expert knowledge, extracted from data, or developed using a hybrid approach that combines both methods.

Note that each part of a fuzzy rule involves fuzzy concepts as follows:
\begin{itemize}
	\item
\textbf{The ``IF" part (antecedent)}: It defines the conditions using fuzzy sets and linguistic variables. For example, ``temperature is high" where ``high" is a fuzzy set.
\item
\textbf{The ``THEN" part (consequent)}: It specifies the output or action, again using fuzzy sets and linguistic variables.
\end{itemize}
A fuzzy rule-based system contains the following components:
\begin{itemize}
	\item 
\textbf{Fuzzification}: The first step is converting crisp (exact) input values into fuzzy values based on some membership functions. This process is known as fuzzification. For instance, see Figure \ref{fuzzysubsets}, for 8 Guassian fuzzy subsets of a variable, called sodium intake. 
\item
\textbf{Rule Evaluation}: Fuzzy rules are evaluated based on the fuzzified inputs. This involves applying logical operations (such as AND, OR, NOT) to determine the degree to which each rule applies. For instance, if we have a rule such as ``If \(x\) is low and \(y\) is high, then \(z\) is medium", then by Mamdani's point of view, we can interpret this rule as defining a specific area on the graph representing ``medium'' \(z\). This area is the set of all points whose membership degrees are at most the minimum of \(\mu_{low}(x)\) and \(\mu_{high}(y)\).

\item
\textbf{Aggregation}: The outputs of individual rules are combined to form a single fuzzy set. This step involves merging the results from multiple rules. For instance,  In Mamdani's approach to fuzzy logic, the aggregated fuzzy output is obtained by taking the union of regions resulting from the evaluation of each rule.
\item
\textbf{Defuzzification}: The final step is converting the aggregated fuzzy output set into a crisp output value. There are several methods for defuzzification as follows:
\begin{itemize}
	\item Centroid Method, which calculates the center of gravity of the obtained fuzzy set.   The  $x$-coordinate  of the centroid is the output value.
	\item Bisector Method, which finds the vertical line that divides the area under the fuzzy set curve into two equal halves. The $x$-coordinate of this line is the output value.
	\item 
	Mean of Maximum (MoM), which  averages the $x$-coordinates of the elements of the fuzzy set with the highest membership degrees. 
	\item Smallest of Maximum (SoM), which  takes the minimum of the $x$-coordinates of the elements of the fuzzy set with the highest membership degrees. 
	\item Largest of Maximum (LoM), which  takes the maximum of the $x$-coordinates of the elements of the fuzzy set with the highest membership degrees. 
\end{itemize}
\end{itemize}
To learn more about fuzzy rule-based systems, refer to \cite{driankov2013introduction, wang1994adaptive}. 
\subsection{Causal Inference}\label{Causal Inference}
Causal inference is a process used in statistical analysis and research to determine whether a cause-and-effect relationship exists between variables. It goes beyond establishing correlations or associations by trying to discern whether a change in one variable causes a change in another. The main approaches in causal inference are around two frameworks: Pearl and Rubin-Neyman  causal frameworks. The first one is based on the concept of causal diagrams and do-calculus. Pearl approach uses DAGs to represent causal relationships among variables. These graphs help identify and control for confounding variables\footnote{In causal inference, a confounder is a variable that influences both the treatment and the outcome, potentially leading to a spurious association between them}, enabling the estimation of causal effects even from observational data. The key contribution of Pearl framework is the introduction of do-calculus, a set of rules for manipulating DAGs to determine the effect of interventions. On the other hand, the Rubin-Neyman causal framework focuses on the concept of counterfactuals — what would have happened to an individual or a group if they had received a different treatment or been in a different condition. The Rubin-Neyman model defines the causal effect as the difference between potential outcomes under different treatment conditions. A critical aspect of this framework is the focus on the assignment mechanism, especially in observational studies, to approximate random assignment and estimate causal effects.  

Below, we provide a concise overview of the notations and causality concepts addressed in this paper.
Let $Y_i(t)$ denote the potential outcome of unit $i$ under treatment $t$, where $t$ can take values 0 or 1, representing the control and treatment conditions, respectively. 
	The Average Treatment Effect (ATE)  is defined as the difference in the expected outcomes between the treated and control groups:
$\ATE = \E[Y(1) - Y(0)]$.
The limitation of ATE often is referred to as the counterfactual problem, the fundamental problem of causal inference, or the problem of missing data. Indeed,  this problem lies in the impossibility of observing both potential outcomes for a single unit at the same time, which  makes it challenging to directly measure the causal effect of the treatment.   The consistency assumption specifies that if an individual receives a particular level of treatment, then the observed outcome $Y$ for that individual is exactly the potential outcome corresponding to that level of treatment ($Y(t) = Y$, if $T=t$).
	The ignorability assumption states that the potential outcomes are independent of treatment assignment:
		$(Y(0), Y(1)) \perp T$.
This assumption is also known as the unconfoundedness or no hidden bias assumption.
	If ignorability and consistency assumptions are satisfied, then one could see that $\ATE = \E[Y|T=1]-\E[Y|T=0]$. 
	Conditional ignorability is another assumption that holds when the potential outcomes are independent of treatment assignment for any value of covariates $X$:
	$(Y(0), Y(1)) \perp T \mid X=x$ for all $x$.
	Similar to the ignorability assumption, here  if the  conditional ignorability and consistency assumptions are satisfied, then one could see that $\ATE = \E_X\bigl[\E[Y\mid T=1, X=x]-\E[Y\mid T=0, X=0]\bigr]$. 
	These notations and assumptions are fundamental in the analysis of causal effects, enabling researchers to estimate the average treatment effect under certain conditions, using statistical methods that rely on observed data. Ignorability and conditional ignorability are posited as methods for achieving the identifiability of ATE. Identifiability comprises a series of assumptions which, when met, permit the estimation of a causal effect formula through observed data and statistical techniques.
	
	When dealing with non-binary treatments, there are several approaches to analyzing their effects on outcomes. One method is to binarize the treatment, simplifying the analysis to a comparison between two distinct groups. Alternatively, the ATE for continuous treatment levels can be articulated more precisely. For a given treatment level \(t\), the ATE can be defined as
		$\ATE(t) = \E[Y(t+1) - Y(t)]$,
	which represents the expected change in the outcome when the treatment level is incrementally increased by one unit. Moreover,  the effect of the treatment can be modeled as the derivative of the expected outcome with respect to the treatment level, expressed as
		$\partial \E(Y(t))/\partial t$. 
	This formulation estimates the instantaneous effect of a small change in the treatment level on the expected outcome, suitable for continuous treatment variables.
	
	\subsection{Structural Equation Modeling}
	
	Structural Equation Models (SEMs) are comprehensive statistical approachs that model relationships among multiple variables, both observed and latent. It incorporates aspects of multiple regression and factor analysis, allowing for the analysis of complex causal relationships.
For instance, 	consider a model examining how student performance $Y$ is influenced by teaching quality $X_1$, student motivation $X_2$, and study hours $X_3$:
	\begin{align*}
		Y &= \beta_1 X_1 + \beta_2 X_2 + \epsilon_1, \\
		X_1 &= \gamma_1 X_3 + \epsilon_2, \\
		X_2 &= \gamma_2 X_3 + \epsilon_3
	\end{align*}
	where $\epsilon_1$, $\epsilon_2$, and $\epsilon_3$ are error terms.

 	SEMs facilitate the identification and estimation of direct and indirect effects within a hypothesized model, making it a powerful tool for testing theories about causal relationships under the assumption that the model is correctly specified.

\subsection{Probabilistic Fuzzy Logic}\label{Probabilistic Fuzzy Logic}
In this research, we implement a distinct probabilistic approach grounded in fuzzy logic principles, as outlined in  \cite{saki2022fundamental}. Consider \(T\) as a stochastic variable, with \(F\) symbolizing a fuzzy attribute within the value domain of \(T\). Our approach is structured through two phases:

\begin{enumerate}
	\item Initially, we randomly choose a conceivable value for \(T\).
	\item With \(T = t\) determined from the initial phase, we engage in a Bernoulli trial with potential outcomes:
	$\{\text{Selecting \(t\) as \(F\)},  \text{Not selecting  \(t\) as \(F\)}\}$.
\end{enumerate}
The above two phases are representative of a stochastic variable \(\xi_{T,F}\), and the result from the latter phase is expressed through \(\xi_{t,F}\). Here, these variables assume the value \(t_F\), denoting the event ‘Not selecting \(t\) as \(F\)’, where \(t_F\) is a potential but improbable value of \(T\) in the context of the second phase, serving as a marker for non-endorsement. The above experiment, including two phases, is referred to as Experiment (\(\star\)) in our discourse. As an illustration, if \(F = \text{high}\), and in the second phase of Experiment (\(\star\)), the likelihood of ‘Selecting \(t\) as \text{high}' is represented by \(\mu_{\text{high}}(t)\), then any point \(t\) in the range \([0.5,1]\) may be identified as \(t_{\text{high}}\), because it is associated with a membership degree of 0 concerning ‘high’.


\section{Main Results}\label{Main Results}
In this section, we first introduce a generalization of the ATE formula. This generalization allows for interventions where the treatment variable is assigned to individuals based on a \underline{specified distribution}, rather than fixing the treatment level directly. Following this, we define the FATE as a special case of this generalized ATE, wherein the distribution enforced upon the treatment variable incorporates fuzzy concepts. We then proceed to examine the FATE, comparing it to the classical ATE to highlight differences and similarities. Additionally, we develop a fuzzy version of the ignorability assumption, along with its implications, to facilitate the identification of the FATE formula. Furthermore, we explore the stability of the FATE formula when treatment values are perturbed in relation to fuzzy attributes. 
\subsection{Generalized Average Tratment Effect }
Assume we have a DAG or an SEM and we need to determine the causal effect of a finite discrete treatment variable \( T \) on an outcome variable \( Y \). Consider a scenario where we actively manipulate the treatment assignment to follow a uniform probability distribution, denoting this as \( T^U \). For non-dichotomous treatment variables, it is typical to define a benchmark threshold \( T_0 \). Treatment values greater than \( T_0 \) are recoded as 1, and those below as 0. In this setup, we can intervene to uniformly assign treatment values \( t \geq T_0 \). However, this does not guarantee that the average outcome is well-defined, because repeating this assignment could lead to different outcomes. To address this, we consider all possible assignment scenarios where treatment values \( t \geq T_0 \) are uniformly assigned and average these outcomes. Assume that $m$ is  the minimum value of $T$, and $T_{\ge T_0}^U$  is the discrete uniform distributions on the values $t$ of $T$ with  $t\ge T_0$. This approach gives us \( \mathbb{E}_{t\sim T_{\ge T_0}^U}\left[\mathbb{E}[Y(t) ]\right]
 \). Similarly, for cases where the treatment values  \( t \leq T_0 \) are uniformly assigned, we compute \( \mathbb{E}_{t\sim T_{\le T_0}^U}\left[\mathbb{E}[Y(t) ]\right]
 \), where  $T_{\le T_0}^U$ is the uniform distribution on the values $t$ of $T$ with $t\le T_0$. The causal effect can then be defined by the formula:
\[
\text{ATE}_{T_0} ^U:= \mathbb{E}_{t\sim T_{\ge T_0}^U}\left[\mathbb{E}[Y(t) ]\right]
- \mathbb{E}_{t\sim T_{\le T_0}^U}\left[\mathbb{E}[Y(t) ]\right].
\]
To accurately define the potential outcomes $Y(T_{\le T_0}^U=t)$ and $Y(T_{\ge T_0}^U=t)$ and consequently $\ATE_{T_0}^U$, it is necessary to employ the \uline{Stable Unit Treatment Value Assumption (SUTVA)} as described within the Rubin-Neyman causal framework. In essence, SUTVA allows researchers to isolate the effect of a treatment, ensuring that variations in outcomes can be attributed solely to the treatment itself rather than to differences in how it was administered or interactions between units. 
The SUTVA consists of two key principles:
\begin{enumerate}
	\item \textbf{No Interference Between Units:} The outcome observed on one unit (e.g., a person, an animal, a school, etc.) should be unaffected by the treatment assigned to any other unit. This principle ensures that the treatment of one individual does not influence the outcomes of others.
	\item \textbf{Consistency:} The treatment applied to any unit must be consistent across all units receiving it. This means that the treatment is administered in the same way and has the same effect wherever it is applied.
\end{enumerate}
We assume that SUTVA holds for both $T_{\le T_0}^U$ and $T_{\ge T_0}^U$. 
Now,  to calculate $\ATE_{T_0}$, let observe what happens if we have a version of the ignorability and the consistency assumptions such as the following:
\begin{itemize}
	\item Uniform \textit{\(T_0\)-Ignorability Assumption}:  \(Y(T_{\le T_0}^U =t)\) and \(Y(T_{\ge T_0}^U =t)\) are independent from \(T_{\le T_0}^U\) and \(T_{\ge T_0}^U\), respectively (in the scenarios such as above where the treatment is assigned entirely at random, this assumption is met).
	\item \textit{Uniform \(T_0\)-Consistency Assumption}: For an individual receiving a treatment level \(T_{\le T_0}^U=t\) or \(T_{\ge T_0}^U=t\),  the observed outcome \(Y\) is equivalent to the potential outcome (this assumption clearly connects the theoretical variables of potential outcomes with the practical variables of observed outcomes).
\end{itemize}
Then, by the above two assumptions, we note  that $\E[Y(t)] = \E[Y\mid t]$, and hence
\begin{align*}
	\text{ATE}_{T_0} ^U&:=\mathbb{E}_{t\sim T_{\ge T_0}^U}\left[\mathbb{E}[Y\mid t ]\right]
	- \mathbb{E}_{t\sim T_{\le T_0}^U}\left[\mathbb{E}[Y\mid t ]\right].
\end{align*}
Assume that the probabilities of assigning a treatment value by $T_{\le T_0}^U$ and $T_{\ge T_0}^U$ to an individual are $p$ and $q$, respectively. Then, we have that 
\[\text{ATE}_{T_0}^U= q\left(\sum_{t\ge T_0}\E[Y\mid t]\right) - p\left(\sum_{t\le T_0}\E[Y\mid t]\right)\]
Assuming \(T_0\) is chosen the middle point of the value set of $T$, implies a significant symmetry in the distribution of \(T^U\) around \(T_0\).  This symmetry is `often' crucial as it underlines the balanced nature of the distribution with respect to \(T_0\).  Then,  $q=p$, and hence
\[\text{ATE}_{T_0}^U= p\left(\sum_{t\ge T_0}\E[Y\mid t]- \sum_{t\le T_0}\E[Y\mid t]\right).\]

Note that instead of a uniform assignment of treatment, an arbitrary intervention on the distribution of \(T\) could be implemented. For instance, consider assigning \(T\) to individuals based on two given  distributions gives us two new treatment variables denoted by  \(T_P\) and \(T_Q\). In this context, one could establish the following formula:
\[
\ATE_{P}^Q(Y; T) := \mathbb{E}_{t\sim T_Q}[\E[Y(t)]] - \mathbb{E}_{t\sim T_P}[\E[Y(t)]].
\]

Assume that the assignment of the treatment \(T_P\) satisfies SUTVA. Consequently, we state that SUTVA is fulfilled with respect to \(T_P\). This ensures that the potential outcome \(Y(T_P=t)\) is well-defined. For a well-defined \(\text{ATE}_P^Q\), it is necessary to assume that SUTVA is fulfilled with respect to both  \(T_P\) and \(T_Q\). In the following remark, we will show how \(Y(T_P=t)\) and \(Y(T_Q=t)\) can indeed differ for a specific value \(t\) of the treatment.

\begin{Remark}
	Observe that the outcomes $Y(T_P=t)$ and $Y(T_Q=t)$ differ significantly. For instance, for a binary treatment $T$, the classical ATE is defined as $\E[Y(1)] - \E[Y(0)]$ (refer to Section \ref{Causal Inference}). Now, we can assume that $T_P$ invariably takes the value $0$, while $T_Q$ invariably takes the value $1$. Therefore, $Y(T_P=1)$ represents an invalid outcome, whereas $Y(T_Q=1)$ does not. For another example, in a medical study, consider a binary treatment where individuals are more likely to receive \( T=1 \) under the assignment of \( T_Q \) than under \( T_P \). Consequently, this may lead to an increase in knowledge about self-care among individuals when \( T_Q \) is applied, potentially affecting the outcome \( Y(T_Q=t) \).
\end{Remark}
If SUTVA is fulfilled  irrespective of the assigned treatment distribution, we refer to this condition as absolute SUTVA. Under absolute SUTVA, the potential outcomes \( Y(T_P=t) \) and \( Y(T_Q=t) \) are identical, allowing us to represent both by \( Y(t) \). \ul{For the rest of this paper, we assume that the absolute SUTVA is fulfilled}.
\begin{Remark}
In the scenario where $T$ is a continuous random variable taking values in the interval $[a, b]$, then 
	\[\ATE_{T_0}^U= q\int_{T_0}^{b}\E[Y\mid t]\,\d t - p\int_{a}^{T_0}\E[Y\mid t]\,\d t,\]
	where $p$ and $q$ are equal to $1/(T_0-a)$ and $1/(b-T_0)$, respectively.
\end{Remark}
\subsection{Fuzzy Average Treatment Effect (FATE) and its Identifiability Criteria}

Let $A$ and $B$ be two fuzzy attributes of $T$. Assume that $\mathbb{P}_A$ (respectively, $\mathbb{P}_B$) represents the probability mass function associated with the random experiment of selecting a particular treatment value as $A$ (respectively, $B$) as stated in the second step of Experiment~($\star$) (see Section \ref{Probabilistic Fuzzy Logic}). We then define the fuzzy average treatment effect (FATE) of $T$ on $Y$ with respect to $(A,B)$ as follows:
\[\text{FATE}_A^B(Y;T) := \text{ATE}_{\P_A}^{\P_B}(Y;T).\]
For simplicity, we denote $T_{\P_F}$ by $T_F$ for any fuzzy attribute $F$. Clearly, our definition of FATE can be used for the continuopus case as well. Now, we can formulate versions of the ignorability and consistency assumptions for any fuzzy attribute $F$ of $T$ as follows:
\begin{itemize}
	\item \textit{$F$-Fuzzy Ignorability Assumption}: $Y(T_F=t)$ and $T_{F}$ are independent for any $t$.
	\item \textit{$F$-Fuzzy Consistency Assumption}: For an individual receiving a treatment level $T_{F} = t$, the observed outcome $Y$ is equivalent to the potential outcome, denoted as $Y(T_F=t) = Y$.
\end{itemize}
Now assume that  the above two assumptions are satisfied for the fuzzy attributes $A$ and $B$ of $T$. Then, we have $\E[Y(t)]= \E[Y\mid t]$, and hence
\[\text{FATE}_A^B(Y;T) = \mathbb{E}_{t\sim T_B}[Y\mid t] - \mathbb{E}_{t\sim T_A}[Y\mid t],\]
which does not suffer from the issue of missing values and can be determined by having a dataset.
\begin{Remark}
We can similarly develop the conditional versions of the $F$-Fuzzy Ignorability and Consistency Assumptions and derive the following expression in the presence of a confounder $C$:
\[\FATE_A^B(Y;T) = \mathbb{E}_C\biggl[ \mathbb{E}_{t\sim T_B\mid C}\bigl[\E[Y\mid t,c]\bigr]\biggr] - \mathbb{E}_C\biggl[ \mathbb{E}_{t\sim T_A\mid C}\bigl[\E[Y\mid t,c]\bigr]\biggr].\]
\end{Remark}
Therfore, we have the following theorem. 
\begin{Theorem}\label{iden}
	$\FATE$ is identifiable under the (conditional) ignorability and the consistency assumptions. 
\end{Theorem}
 \begin{Remark}
 We can also reinterpret the preceding context through Pearl's perspective. In this framework, our generalized ATE can be articulated as follows:
 \[\ATE_P^Q(Y;T) := \mathbb{E}_{t\sim T_Q}[\E[Y|\Do(T_Q = t)] ]-\mathbb{E}_{t\sim T_P}[\E[Y|\Do(T_P = t)] ],\]
 where the \text{do}-operator specifies an intervention in the causal model, setting the treatment variable $T_F$ to the level $t$ for $F\in\{A,B\}$. Identifiability in this setting hinges on the \textit{sufficiency assumption}, which posits that all pertinent variables are included and accurately represented in the problem. This assumption ensures that the DAG illustrating the causal relationships among the variables is comprehensive and well-specified. The sufficiency assumption is crucial as it underpins the ability to draw valid causal inferences from the model.
 
 \end{Remark}
\subsection{Analyzing Results on FATE and Its Comparison with ATE }
We note that our generalized ATE formula measures the average (causal) changes of $Y$ with respect to the interventional changes of $T$. In the contrast, the classic versions of   ATE formula measure the change of $Y$ while interventionally increasing the value of $T$ by 1 or a an infinitesimal value of $T$ (see Section \ref{Causal Inference}). Hence, to compare our formula to the classic versions of ATE, we suggest dividing the above value by $\E[T_Q - T_P]$. Therefore, we can define the normalized version of our generalized ATE as follows:
\[\NATE_P^Q(Y;T):= \frac{\ATE_P^Q(Y;T)}{\E[T_Q-T_P]},\]
and hence 
\[\NFATE_A^B(Y;T):=\frac{\FATE_A^B(Y;T)}{\E[T_Q-T_P]}.\]
\begin{Remark}
	We can go further and for any positive integer $d$, we divide the $\ATE_P^Q(Y;T)$  by $\E[T_Q^d - T_P^d]$ to normalize $\FATE $ with respect to the changes of higher degrees. Thus, we define
	\[\NATE_P^Q(Y;T)(d):= \frac{\mathrm{ATE}_P^Q(Y;T)}{\E[T^d-T'^d]}. \]
\end{Remark}

In the following example, we explore the concept of causality through the lens of ATE  given an SEM. Further along in this paper, we will delve into this example employing our FATE formula.
\begin{Example}\label{linear}
Consider the following  SEM, and assume that \( U_T \) and \( U_Y \) are independent. The model is given by:
\begin{align*}
	X &= U_X, \\
	T &= \alpha X + U_T, \\
	Y &= \beta T^n + \gamma X + U_Y,
\end{align*}
where \( n \) is a positive integer. Initially, let \( n = 1 \). In this scenario, the $\ATE$ of \( T \) on \( Y \) is \( \beta \), given that
\[\E[Y(T=1)] -  \E[Y(T=0)] = \left( \beta + \gamma\E[X] + \E[U_Y]\right) - \left(\gamma\E[X]+ \E[U_Y]\right) = \beta.\]
Note: If there is a correlation between \( U_Y \) and \( U_T \), then the $\ATE$ of \( T \) on \( Y \) is not necessarily equal to \( \beta \). The exact value of $\ATE$ depends on the complex relationships between the endogenous variables and the unobserved error terms, such as \( U_Y \) and \( U_T \).

Next, consider \( n = 2 \). In this case, there are two different classic interpretations for the $\ATE$ of \( T \) on \( Y \):
\begin{enumerate}
	\item By keeping \( X \) constant and increasing \( T \) from \( t \) to \( t + 1 \), the change in \( Y \) is observed, yielding \( \beta(2t + 1) \), which depends on \( t \).
	\item If \( T \) is a continuous random variable, and \( X \) is held constant, then increase \( T \) from \( t \) to \( t + h \), where \( h \) is small and positive. The result is then divided by \( h \) to make it comparable across different small values of \( h \). Assuming \( h \to 0^+ \), the limit yields:
	\[\lim_{h \to 0^{+}} \frac{\beta(t + h)^2 - \beta t^2}{h} = \frac{\d}{\d t}(\beta t^2) = 2\beta t.\]
\end{enumerate}

Therefore, for $n=2$, the difference between the above two approaches is a constant $\beta$.  It's noteworthy that for \( n > 2 \), these approaches yield divergent results. However, the partial derivative of \( \beta t^n \) (second approach) is always the leading term of \( \beta(t + 1)^n - \beta t^n \) (first approach).
\end{Example}
 Assume that $T$ is a continuous random variable and $(A,B)$ is a pair of fuzzy attributes of $T$. In general,  the probability density functions $f_A$ and $f_B$  migh not depend on $\mu_A$ and $\mu_B$. However, one could subjectively use the following definitions to select values of $T$ as $A$ or $B$:
\[f_A(t) =\frac{\mu_{A}(t)}{\lVert A\rVert},\qquad f_B(t) =\frac{\mu_{B}(t)}{\lVert B\rVert},\quad \lVert F\rVert:=\int_{-\infty}^{\infty} \mu_F(u)\,\d u,\quad F\in\{A,B\}. \]
The \ul{\textit{standard model}} for $f_A$ and $f_B$ will be established as the one where both $f_A$ and $f_B$ are defined in the aforementioned  manner. Similarly, we can define this standard model in the discrete case. 
In the remainder of this paper, we focus exclusively on the continuous case, although similar results apply to the discrete case as well.
\begin{Example}
	Let us assume that we are given the SEM of Example \ref{linear} for $n=1$. Let $T$ be a continuous random variable  with $T\in[0,L]$. We can consider different definitions for $\mu_l = \mu_{low}$ and $\mu_h = \mu_{high}$ as it is shown in Figure \ref{fatedifferentmembershipfunctions}. In each case, by considering the standard model for $f_l$ and $f_h$, the $\FATE_l^h(Y\; T)$  has been calculated. 
	\begin{figure}
		\centering
		\subfloat[$\mathrm{FATE} = \frac{2L}{3}\beta$\label{simple}]{
			\begin{tikzpicture}
				\draw[->] (0,0) -- (4.5,0) node[right] {$t$};
				\draw[->] (0,0) -- (0,2) node[above] {$\mu(t)$};
				\draw[thick, blue] (0,1.5) -- node[above right]{low} (2,0)node[black, below]{$\frac{L}{2}$};
				\draw[thick, red] (2,0) --(4,1.5);
				\draw[red] (3,0.68)  node[above left]{high};
				\draw[thick, green, dashed] (4, 1.5) -- (0,1.5) node[black, left]{$1$};
				\draw[thick, green, dashed] (4, 1.5) -- (4,0) node[black, below]{$L$};
		\end{tikzpicture}}
		\subfloat[$\mathrm{FATE}_l^h= \beta\left(L - \frac{2M}{3}\right)$]{
			\begin{tikzpicture}
				\draw[->] (0,0) -- (4.5,0) node[right] {$t$};
				\draw[->] (0,0) -- (0,2) node[above] {$\mu(t)$};
				\draw[thick, blue] (0,1.5) -- node[left]{low} (3,0)node[black, below]{$M$};
				\draw[thick, red] node[black, below right]{$L-M$}(1,0) -- (4,1.5);
				\draw[red] (3.6,0.33)  node[above left]{high};
				\draw[thick, green, dashed] (4, 1.5) -- (0,1.5) node[black, left]{$1$};
				\draw[thick, green, dashed] (4, 1.5) -- (4,0) node[black, below]{$L$};
			\end{tikzpicture}
		}\\
		\subfloat[$\mathrm{FATE}_l^h = \frac{11L}{18}\beta$]{
			\begin{tikzpicture}
				\draw[->] (0,0) -- (4.5,0) node[right] {$t$};
				\draw[->] (0,0) -- (0,2) node[above] {$\mu(t)$};
				\draw[thick, blue] (0,1.5)	node[black, left]{$1$} -- node[above right]{low} (1,1.5) --(2,0)node[black, below]{$\frac{L}{2}$};
				\draw[thick, red] (2,0) --(3,1.5)--(4,1.5);
				\draw[red] (3.5,1.4)  node[above left]{high};
				\draw[thick, green, dashed] (3, 1.5) -- (1,1.5);
				\draw[thick, green, dashed] (4, 1.5) -- (4,0) node[black, below]{$L$};
				\draw[thick, green, dashed] (3, 1.5) -- (3,0) node[black, below]{$\frac{3L}{4}$};
				\draw[thick, green, dashed] (1, 1.5) -- (1,0) node[black, below]{$\frac{L}{4}$};
		\end{tikzpicture}}
		\subfloat[$\mathrm{FATE}_l^h \approx \frac{6L}{10}\beta$,  ($\mathrm{std }=\frac{L}{4}$)]{
			\begin{tikzpicture}
				\begin{axis}[scale =0.5,
					xmin=0, xmax=4.3,
					ymin=0, ymax=1.5,
					ticks=none,
					width=10cm, 
					height=5cm, 
					hide x axis, 
					hide y axis,
					]
					
					\addplot [
					domain=0:4, 
					samples=100, 
					color=blue,
					thick
					]
					{1.3*exp(-0.5*(x^2)/0.5)};
					
					\addplot [
					domain=0:4, 
					samples=100, 
					color=red,
					thick
					]
					{1.3*exp(-0.5*((x-4)^2)/0.5)};
					
				\end{axis}
				\draw[->] (0,0) -- (4.5,0) node[right] {$t$};
				\draw[->] (0,0) -- (0,2) node[above] {$\mu(t)$};
				\draw(0,1.5) node[left] {$1$};
				\draw (2,0) node[below] {$\frac{L}{2}$};
				\draw[thick, dashed,green] (0,1.5) -- (4,1.5);
				\draw[thick, dashed,green] (4,1.5) -- (4,0)node[below, black] {$L$};
				\draw[red] (3.3,0.8)  node[above left]{high};
				\draw[blue] (1.6,0.85)  node[above left]{low};
			\end{tikzpicture}
			
		}
		\caption{Four different types of definitions for each of $\mu_{low}$ and $\mu_{high}$. In each case, the $\FATE_{low}^{high}(Y;T) $ has been calculated. }\label{fatedifferentmembershipfunctions}
	\end{figure}
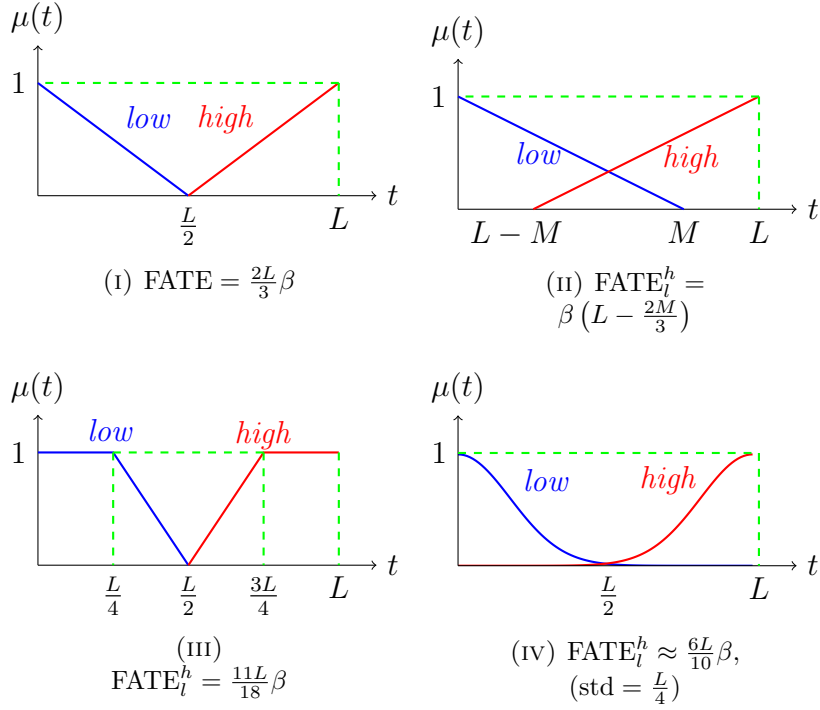
\end{Example}

Let $T$ be a random variable. As shown in Figure \ref{fatedifferentmembershipfunctions}, for every value of $T=t$, it holds true that $\mu_h(t) = \mu_l(L-t)$. This relationship stems from the property that the graph of $\mu_l$ is essentially a mirror image of the graph of $\mu_h$, reflected along the line $T=L/2$. It follows that in the standard model for $f_A$ and $f_B$, we have that $f_B(t) = f_A(L-t)$. In general, for instance when $f_A$ and $f_B$ are not necessarily comming from the satndard model, we call the pair $(A,B)$ of fuzzy attributes \ul{\textit{symmetric}} if $f_B(t) = f_A(L-t)$ for any $T=t$. 

In the subsequent theorem, we elucidate the connection between  NFATE and ATE when a confounder is present. 
We note that in this theorem, the simplified assumption that $T \in [0, L]$ is not necessary, and $T$ can take values from any interval.
\begin{Theorem}\label{trianglemain}
	Assume that we are given the following SEM with independent $U_T$ and $U_Y$:
	\begin{align*}
		X&=U_X\\
		T&=h(X) +U_T\\
		Y&=g(T,X) + U_Y,
	\end{align*}
	where $g$ is a polynomial of degree $r$. Then, for any pair $(A,B)$ of fuzzy attributes with $\E(T_A)\neq \E(T_B)$:
	\begin{enumerate}
		\item For $r=1$,  we have that $\NFATE_A^B(Y;T) =\beta = \ATE(Y;T)$.
		\item For $r=2$,  if $T\in[0,L]$, and $(l,h)=(low, high)$ is a pair of symmetric fuzzy attributes of $[0,L]$, then 
		\[	\NFATE_l^h(Y;T) = \text{ The mean value of $\ATE(Y;T)$ on $[0,L]$},\]
		where we have considered the partial derivative point of view of $\ATE(Y;T)$. 
		\item In general, we have that
		\[\NFATE_A^B(Y;T) = \beta \left(\frac{\E_{t\sim T_B}\bigl[\E[g(t,X)]\bigr]- \E_{t\sim T_A}\bigl[\E[g(t,X)]\bigr]}{\E[T_B] - \E[T_A]}\right).\]
	\end{enumerate}
\end{Theorem}
\begin{proof}
	(1) Let $g(T,X) = \beta T+\gamma X$. We note that $\E[Y(t)] = \beta t +R$, where $R=\gamma\E[X]+\E[U_Y]$. It follows  that 
	\[\FATE_A^B(Y;T) =\beta\left( \int_{-\infty}^{\infty}tf_B(t)\,\d t - \int_{-\infty}^{\infty}tf_A(t)\,\d t\right) =\beta\left( \E[T_B]- \E[T_A]\right),\]
	while $\E[T_B]- \E[T_A]$ is the same as $\E(T_B-T_A)$. Therefore, 
	\[\NFATE_A^B(Y;T) =\beta = \ATE(Y;T).\]
	
	(2) Since $(l,h)=(low, high)$ is a pair of symmetric  fuzzy attributes of $[0,L]$, for any $t\in [0,L]$, we have that $f_h(t) = f_l(L-t)$. Let $g(T,X) = \beta T^2+\eta TX+ \zeta X^2+\delta X$. We note that $\E[Y(t)] = \beta t^2+\eta t\E[X] +R'$, $R'= \zeta\E[X^2] +\delta\E[X] +\E[U_Y]$. It follows 
	\begin{align*}
		\E_{t\sim T_h}[\E[Y(t)]]&= \int_{0}^L \E[Y(t)]f_h(t)\,\d t  =\beta\int_0^Lt^2f_l(L-t)\,\d t \\
		&+\eta\E[X]\int_{0}^Ltf_l(L-t)\,\d t + R'\int_{0}^Lf_h(t)\,\d t\\
		&= \beta\int_0^L (L-t)^2f_l(t)\,\d t +\eta\E[X]\int_0^L (L-t)f_l(t)\,\d t +R'.
	\end{align*}
	Also, we have that
	\[\E_{t\sim T_l}[\E[Y(t)]]=  \beta\int_0^L t^2f_l(t)\,\d t +\eta\E(X)\int_0^L tf_l(t)\,\d t +R'.\]
	Thus, 
	\begin{align*}
		\FATE_l^h(Y;T) &= \beta\left(\int_0^L(L-t)^2f_l(t)\,\d t - \int_0^Lt^2f_l(t)\,\d t\right)\\
		&+\eta\E[X]\left(\int_0^L(L-t)f_l(t)\,\d t - \int_0^Ltf_l(t)\,\d t\right)\\
		&= \beta\left(L^2\int_0^Lf_l(t)\,\d t - 2L\int_0^Ltf_l(t)\,\d t\right)\\
		&+ \eta\E[X]\left(L\int_0^Lf_l(t)\,\d t -2 \int_0^Ltf_l(t)\,\d t\right)\\
		&= \beta\left(L^2 - 2L\E[T_l]\right) +\eta\E[X]\left(L-2\E[T_l]\right).
	\end{align*}
	Similarly, one could see that $\E[T_h-T_l] = L-2\E[T_l]$, which implies that $\NFATE_l^h(Y;T) = \beta L+\eta \E[X]$. It is easy to see that 
	\[\frac{1}{L} \int_0^L \ATE(Y;T)\,\d t = \beta L+\eta \E[X], \]
	which completes the proof.

	(3) One could prove it by a similar argument to the proof of (2).
\end{proof}
\subsection{Stability of FATE}
It's important to recognize that causal problems can often be chaotic, with slight variations in the treatment variable leading to significant fluctuations in the outcome. Nonetheless, in this subsection, we seek to identify a condition that ensure the stability of FATE when subjected to minor adjustments of the treatment variable. 
To achieve this, it is necessary first to establish a metric for the set of fuzzy attributes associated with a given set. To this end, we employ the $L^1$-norm, as detailed in the subsequent discussion.
The $L^1$-norm ($\lVert\,\cdot\,\rVert_{L^1}$) of a continuous real-valued function is defined as the integral of its absolute value over its domain. We can turn the set $\mathcal{F}(S)$  of all fuzzy attributes of a set $S$ into a metric space by defining the distance of $F,F'\in\mathcal{F}(S)$ to be $\d(F,F')= \lVert f_F - f_{F'} \rVert_{L^1}$. Naturally, we can define a metric on $\mathcal{F}(S)\times\mathcal{F}(S)$ as follows:
\[\d((A,B),(A',B'))=\sqrt{\d(A,A')^2+\d(B,B')^2},\quad A,A',B,B'\in\mathcal{F}(S).\]
 By the following definition, we ensure that our causal problem is not chaotic. 
\begin{Definition}\label{stable}
We define our causal problem \ul{stable} under the treatment variable $T$, when $\E[Y(t)]$ is a bounded function of $t$.  
\end{Definition}
In the following theorem, we demonstrate the robustness of FATE with respect to minor variations in fuzzy attributes. 
\begin{Theorem}\label{stability}
$\FATE$ remains  stable (Lipschitz continuous\footnote{Let \( X \) be a metric space and \( f: X \to \mathbb{R} \) be a function. Then \( f \) is said to be Lipschitz continuous if there exists a real constant \( L \geq 0 \) such that for all \( x, y \in X \),
	$
	|f(x) - f(y)| \leq L \d_X(x ,y)8
$.
}) if our causal problem is  stable.
\end{Theorem}
\begin{proof}
	Let $M$ be a bound for $\E[Y(t)]$, and let $\epsilon>0$. Assume that $(A',B')$ is another pair of fuzzy attributes such that $\d((A,B),(A',B'))<\epsilon/(2M)$. Thus, $\lVert f_A - f_{A'}\rVert_{L^1} <\epsilon/(2M)$ and $\lVert f_B - f_{B'}\rVert_{L^1} <\epsilon/(2M)$. Then, if $G(t) = \E[Y(t)]$, we have that 
\begin{align*}
	\left| \FATE_{A'}^{B'}(Y;T) - \FATE_A^B(Y;T)\right|&\le \lVert G(f_A - f_{A'})\rVert_{L^1} + \lVert G(f_B - f_{B'})\rVert_{L^1}\\
	&\le M\left(\lVert f_A - f_{A'}\rVert_{L^1} + \lVert f_B - f_{B'}\rVert_{L^1}\right)\\
	&\le 2M\d((A,B),(A',B'))<\epsilon.
\end{align*}
\end{proof}
\begin{Corollary}\label{kl}
	FATE remains stable (it satisfies Hölder condition\footnote{Let \( X  \) be a metric space and \( f: X \to \mathbb{R} \) be a function. Then we say  \( f \) satisfies the  Hölder condition  if there exists a real constant \( L \geq 0 \) and a positive constant $\alpha$ such that for all \( x, y \in X \),
		$
		|f(x) - f(y)| \leq L \d_X(x ,y)^{\alpha}
		$.
		}), when we consider the  $\kl$ divergence distance on $\mathcal{F}(S)$.
\end{Corollary}
\begin{proof}
It follows directly from Theorem \ref{stability} and the  Pinsker’s inequality, which says  
	\[D_{\kl}(P||Q) \ge \frac{1}{2}\lVert P-Q\rVert_{L^1}^2,\]
	for every two distributions $P$ and $Q$ on the same set. 
\end{proof}
\section{Generalized Fuzzy Average Treatment Effect}\label{Generalized Fuzzy Average Treatment Effect}
In the definition of $\FATE_A^B(Y;T)$, the occurrence frequency of the values of $T$ is not taken into account. Consequently, even if the probability of $T$ taking a particular value $t_0$ is exceedingly low, the potential outcome associated with $T=t_0$ is given equal importance as the potential outcome for a more frequently occurring value $T=t_1$. In certain real-world scenarios, it may be necessary to alter the significance attributed to the occurrence frequency mentioned earlier. For example, in some applications, minor noises might not be very important, whereas in healthcare sciences, a rare symptom could be highly significant. Therefore, we introduce a generalized form of $\FATE$, denoted as $\GFATE$, to accommodate the requirements discussed above. 
This can be \uline{partially} formalized by addressing the following question:
\begin{Question}
	Given a fuzzy attribute $A$, how do we quantify the probability of selecting a value $T=t$ as $A$?
\end{Question}
Following \cite{saki2022fundamental}, the response is given by $\P_{\zeta_{T,A}}(t)$, where $\zeta_{T,A}$ may follow any distribution. Subsequently, we can define the generalized $\FATE$ in a measure-theoretic framework as:
\[\GFATE_A^B(Y;T)= \ATE_{\zeta_{T,A}}^{\zeta_{T,B}}=\E_{t\sim \zeta_{T,B} }[\E[Y(t)]] - \E_{t\sim \zeta_{T,A} }[\E[Y(t)]] .\]
To make this formula comparable to the ATE formula, we normalize it as follows:
\[\NGFATE_A^B(Y;T)=\frac{\GFATE_A^B(Y;T)}{\E[\zeta_{T,B}-\zeta_{T,A}]}.\]
If the condition of being $A$ is independent of selecting $T=t$, then one could define: \[f_{\zeta_{T,A}}(t)=\frac{f_T(t)f_A(t)}{\int_{-\infty}^{\infty}f_T(u)f_A(u)\,\d u}. \] 
The \ul{\textit{independent model}} for $f_A$ and $f_B$ will be established as the one where $f_{\zeta_{T,A}}$ is defined in the aforementioned  manner. Further, the \ul{\textit{standrad independent model}} is given by an independent model in which 
\[f_A(t)=\frac{\mu_A(t)}{\lVert A\rVert},\quad \lVert A\rVert=\int_{-\infty}^{\infty}\mu_A(u)\,\d u.\]
A pair of fuzzy attributes $(A, B)$ associated with a random variable $T$, where $T\in[0,L]$, is called \ul{\textit{generalized symmetric}} if the condition $f_{\zeta_{T,B}}(t) = f_{\zeta_{T,A}}(L - t)$ holds. It's important to note that even if $(A, B)$ is a  pair of symmetric fuzzy attributes for $T$ (meaning $f_A(t) = f_A(L - t)$), we cannot automatically conclude that $(A, B)$ is \textit{generalized symmetric}. This is the case even in the standard independent model, as the distribution $f_T$ does not necessarily have to be symmetric around the line $T = \frac{L}{2}$.

Now, we can generalize Theorem \ref{trianglemain}, 
as follows:
\begin{Theorem}\label{generalizedtrianglelemma}
	Assume that we are given the  SEM of Theorem \ref{trianglemain}. Then, for any pair $(A,B)$ of fuzzy attributes with $\E[\zeta_{T,A}]\neq\E[\zeta_{T,B}]$:
	\begin{enumerate}
		\item For $r=1$,  we have that $\NGFATE_A^B(Y;T) =\beta = \ATE(Y;T)$,
		\item For $r=2$,  if $T\in[0,L]$, and $(low, high)$ is a pair of generalized symmetric fuzzy attributes of $[0,L]$, then 
		\[	\NGFATE_l^h(Y;T) = \text{ The mean value of $\ATE(Y;T)$ on $[0,L]$},\]
		\item In general, we have that
		\[\NGFATE_A^B(Y;T) = \beta \left(\frac{\E_{t\sim \zeta_{T,B}}\bigl[\E[g(t,X]\bigr] - \E_{t\sim \zeta_{T,A}}\bigl[\E[g(t,X)]\bigr]}{\E[\zeta_{T,A}]-\E[\zeta_{T,B}]}\right).\]
	\end{enumerate}
\end{Theorem}
Similar to the approach for FATE, one could define the (conditional) fuzzy ignorability and fuzzy consistency assumptions. Consequently, a result similar to Theorem~\ref{iden} for $\GFATE$ holds as well. Similarly, the results akin to Theorem \ref{stability} and Corollary \ref{kl} are applicable.

\section{Causality in Fuzzy Systems}\label{Causality in Fuzzy Systems}
In the preceding sections, our focus was on exploring the fuzzy causal effects associated with crisp outcomes. However, let's shift our perspective to a situation where the outcome itself is fuzzy. In such instances, we employ a fuzzy control system to estimate the outcome's values. This approach necessitates the formulation of fuzzy rules. 

Let $Y=g(T,X_1,\ldots,X_n) + U_Y$, and  for any $1\le i\le m$, let $A_1,\ldots, A_l$,  $B_1^i,\ldots, B_{l_i}^i$, and $C_1,\ldots, C_p$ be fuzzy attributes of $T$, $X_i$ and $Y$, respectively. Assume that we are given the following fuzzy rules:
\[ R(i,j_1,\ldots,j_n):\text{If } (T\text{ is } A_{i})\,\&\, (X_1\text{ is } B_{j_1}^1)\,\&\,\cdots\,\&\, (X_n\text{ is } B_{j_n}^n)\text{, then } Y\text{ is } C_{ij_1\cdots j_n}, \]
where  $1\le i\le l$,  $1\le j_k\le l_k$, and $C_{ij_1\cdots j_n}\in \{C_1,\ldots, C_p\}$ for any $1\le k\le n$. Let $\mathscr{R}$ be the set of all fuzzy rules considered above. Then, we denote the output of the fuzzy system with the inputs $T, X_1,\ldots, X_n$ by $Y_{\mathscr{R}}$. Now, we can apply our fuzzy causal effect formulas to compute the causal effect of $T$ on $Y$ with respect to the estimation $Y\approx Y_{\mathscr{R}}$. 

The formulation of fuzzy rules typically involves input from an expert. However, in situations where expert guidance is unavailable or the rules are uncertain, and a dataset is provided, there are various methods available for rule extraction from the dataset (refer to \cite{al2014automatic,ductu2017fast,fakhrahmad2007constructing,kim1999efficient,wang1992generating}). Here, `dataset' refers to a collection of data where the input variables are represented by crisp values, while the outcomes are categorized using fuzzy labels such as \textit{low}, \textit{medium}, and \textit{high}. If the outcome variable in our dataset is crisp as well, the method outlined in the previous sections, which employs crisp analysis, can be applied. However, it's also feasible to derive fuzzy rules from this data and utilize a fuzzy control system to recalculate the outcome values. This approach allows for a more nuanced interpretation of the data, even when the outcomes are initially presented in a crisp format.
\subsection{Probabilistic Rules}
When analyzing a real world problem, we often consider only the most observable variables that have significant effects, though there are many other smaller factors at play. Hence, similar conditions for the considered variables  can lead to different outcomes. Therefore, in a fuzzy system, it is conceivable to have rules with identical antecedents but differing consequents, where a specific probability value  is associated to each rule. Assume the fuzzy rules in our system can be represented as:
\[ R(j_0,j_1,\ldots, j_{n+1}):\text{If } (T\text{ is } A_{j_0})\,\&\, (X_1\text{ is } B_{j_1}^1)\,\&\,\cdots\,\&\, (X_n\text{ is } B_{j_n}^n)\text{, then } Y\text{ is } C_{j_{n+1}} \]
These rules are observed with a probability denoted as $p_{j_0\cdots j_{n+1} }$.
The collection of all fuzzy rules sharing the same antecedents as described is denoted by $R(j_0,j_1,\ldots, j_n)$.  Given 
$\mathscr{R}$
 as the set of all these fuzzy rules, and considering that each instance of rule application involves selecting exactly one rule from each $R(j_0,j_1,\ldots, j_n)$ in $\mathscr{R}$, we can express $\E(Y(t))\approx \E[Y_{\mathscr{R}}(t)]$ with
\[ \E[Y_{\mathscr{R}}(t)]= \E_{\mathscr{R}}[\E[Y(t)|\mathscr{R}]] =  \sum\left(\prod_{j_0,\ldots,j_n}p_{j_0\cdots j_{n+1}}\right)\E\left[Y_{\mathscr{R}}(t)|R(j_0,j_1,\ldots, j_{n+1})\right],\]
where  the summation varies over all choices of $R(j_0,\ldots,j_{n+1})\in R(j_0,\ldots, j_n)$, and  $\prod_{j_0,\ldots,j_n}p_{j_0\cdots j_{n+1}} $ represents the probability of a particular configuration of rules. Furthermore, $\E[Y_{\mathscr{R}}(t)|\mathscr{R})$ represents the expected value of the potential outcome for $T=t$, which is determined by the inputs and a specific configuration of fuzzy rules within the set $\mathscr{R}$. 
Finally, with $\E[Y(t)]$ determined, we can compute $\E_{t\sim T_F}[\E[Y(t)]]$ for a fuzzy attribute $F$ of the value domain of $T$. This allows us to calculate the FATE of $T$ on $Y$ concerning a pair of fuzzy attributes.
\section{Examples}\label{Examples}

In this section, we present practical examples drawn from two case studies: 1) the impacts of sodium intake and age on blood pressure, and 2) the influence of food quality on tipping behaviors. The Python codes for these examples can be accessed via this \href{https://github.com/joseffaghihi/Integrating-Fuzzy-Logic-with-Causal-Inference-Enhancing-the-Pearl-and-Neyman-Rubin-Methodologies.git}{GitHub link}.

\subsection{Sodium Intake and Blood Pressure}\label{Sodium Intake and Blood Pressure}
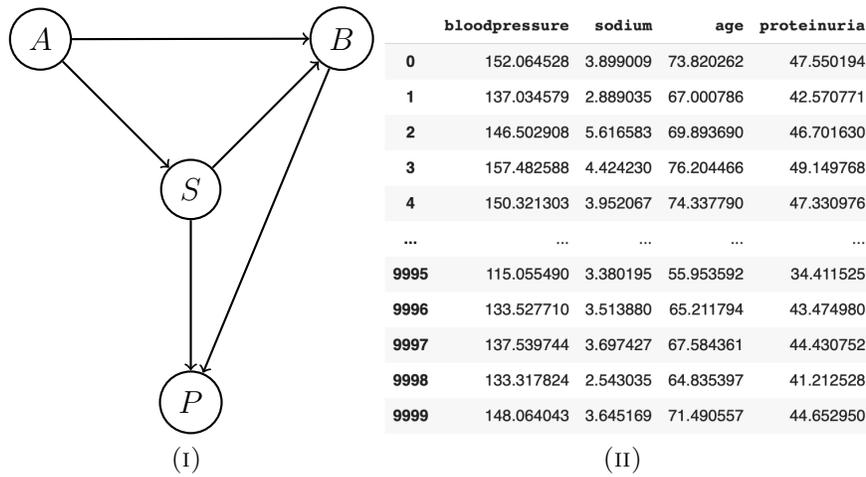
\begin{figure}
		\begin{center}
	\subfloat[\label{sodium}]{
\begin{tikzpicture}[node distance=2cm, auto]
	\node[circle, draw, thick] (S) {$S$};
	\node[circle, draw, thick] (A) [above left=of S] {$A$};
	\node[circle, draw, thick] (B) [above right=of S] {$B$};
	\node[circle, draw, thick] (P) [below=of S] {$P$};
	
	\draw[->, thick] (A) -- (S);
	\draw[->, thick] (A) -- (B);
	\draw[->, thick] (S) -- (B);
	\draw[->, thick] (S) -- (P);
	\draw[->, thick] (B) -- (P);
\end{tikzpicture}}
\subfloat[\label{table-sodium}]{	\includegraphics[scale = 0.43]{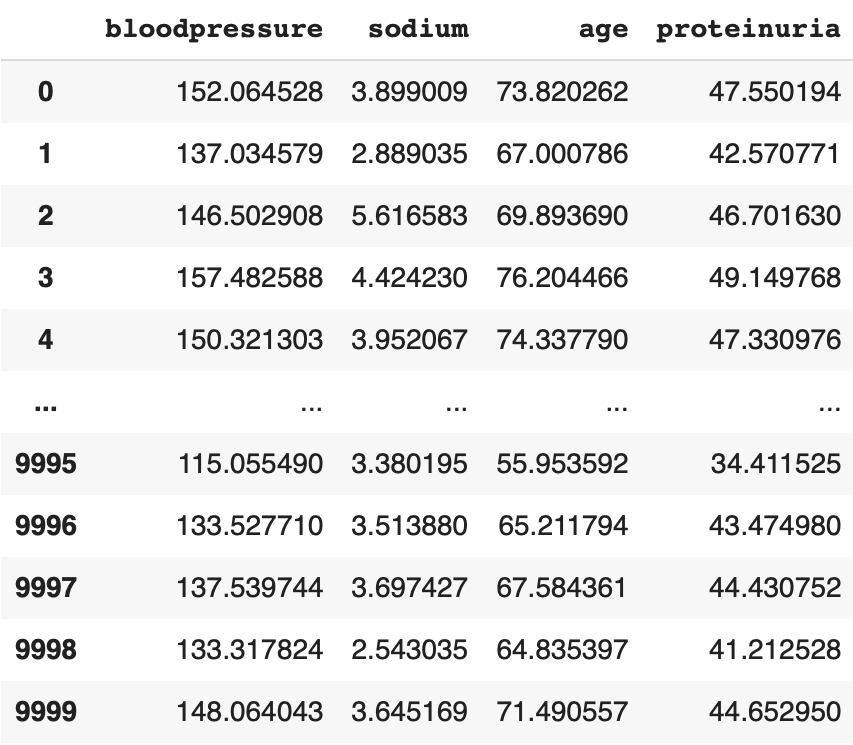}}
		\caption{(I) The DAG describing the relationships between sodium intake ($S$), age ($A$), proteinuria ($P$), and bloodpressure ($B$).\\ (II) The data associated to the DAG shown in (I)}
	\end{center}
\end{figure}

High sodium intake often leads to increased blood pressure, as sodium causes the body to retain water.
Also, older adults may be more sensitive to sodium, potentially leading to a greater risk of high blood pressure with high sodium intake due to decreased sodium excretion.
Blood pressure generally increases with age due to arterial stiffening and plaque build-up, making the effects of sodium on blood pressure more significant in older individuals. Proteinuria, characterized by excess protein in the urine, is often linked to kidney damage. High blood pressure, which can be exacerbated by high sodium intake, can both cause and worsen kidney damage and proteinuria (see \cite{nerbass2018sodium} and \cite{weir2012urinary}). Therefore, one can consider the DAG shown in Figure \ref{sodium}.  

Now, consider the data shown in Figure \ref{table-sodium}. Then the true ATE of sodium intake on blood pressure is 1.05, based on the following data generating process in \cite{luque2019educational}:
\vspace*{0.15cm}
\begin{lstlisting}[caption={Methodology for generating the data presented in Table~\ref{table-sodium} }, label=lst:generate_data]
	def generate_data(n=10000, seed=0, beta1=1.05, alpha1=0.4, alpha2=0.3):
	    np.random.seed(seed)
	    age = np.random.normal(65, 5, n)
   	    sodium = age / 18 + np.random.normal(size=n)
	    bloodpressure = beta1 * sodium + 2 * age + np.random.normal(size=n)
	    proteinuria = alpha1 * sodium + alpha2 * bloodpressure + np.random.normal(size=n)
	    return pd.DataFrame({'bloodpressure': bloodpressure, 'sodium': sodium,
		'age': age, 'proteinuria': proteinuria})
\end{lstlisting}
As demonstrated in  \href{https://github.com/joseffaghihi/fuzzyaveragecausaleffects.git}{github}, by determining the regression line that correlates blood pressure with age and sodium intake, we can maintain a constant sodium intake, set alternately at 1 and then at 0. This approach allows us to calculate the ATE of sodium intake on blood pressure. 

Table \ref{your-label-here} presents a comparison of the results for ATE, FATE, NFATE, GFATE, and NGFATE as follows:

\begin{table}[ht]
	\centering
	\Small
	\begin{tabular}{|c|c|c|c|c|c|c|}
		\hline
		 True ATE, NFATE and NGFATE & True FATE & ATE & FATE  & NFATE &   GFATE & NGFATE \\
		\hline
		1.05 & 5.85 & 1.06 & 5.92 &  1.06 & 2.77 & 1.06\\
		\hline 
	\end{tabular}
	\caption{The true values and the estimated values for the ATE, FATE, NFATE, GFATE, and NGFATE of sodium intake on blood pressure.}
	\label{your-label-here}
\end{table}

The resulting ATE, FATE, NFATE, GFATE and NGFATE values are reasonably accurate approximation of the true effects. The true value of FATE is indicated in Figure \ref{simple}. Additionally, the true values of  NFATE and NGFATE are obtained by Theorem \ref{trianglemain} and Theorem \ref{generalizedtrianglelemma}. 

\subsection{Tipping Problem}
\begin{figure}
\begin{tikzpicture}[node distance=2cm, auto]
	\node[circle, draw, thick] (Q) {Q};
	\node[circle, draw, thick] (S) [right=of Q] {S};
	\node[circle, draw, thick] (T) [below right=of Q] {T};
	
	\draw[->, thick] (Q) -- (T);
	\draw[->, thick] (S) -- (T);
	\draw[->, thick] (S) -- (Q);
\end{tikzpicture}
\caption{The DAG describing the relationships between food quality ($Q$), service quality ($S$), and tip ($T$).}\label{tipdag}
\end{figure}
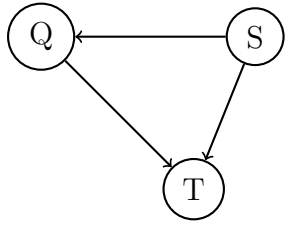

DAG in Figure \ref{tipdag} illustrates the interrelations among food quality, service quality, and the tipping behavior at a restaurant.
To calculate the causal effect of food quality on tipping behavior, we assume the assumptions of conditional ignorability and consistency are met. This leads to the following expression:
\[
\E[T(q)] = \E_C\left[ \E[T(q) \mid c]\right] = \E_C\left[ \E[T(q) \mid q, c]\right] = \E_C\left[ \E[T \mid q, c]\right].
\]
We then generate a sample from $S$ and employ a fuzzy logic system to estimate $\\E[T \mid q, c]$, subsequently averaging these estimates. To proceed, we assume that   \( S \) is normally distributed with a mean of 7 and a variance of 2.

\begin{figure}
	\begin{center}
		\includegraphics[scale = 0.3]{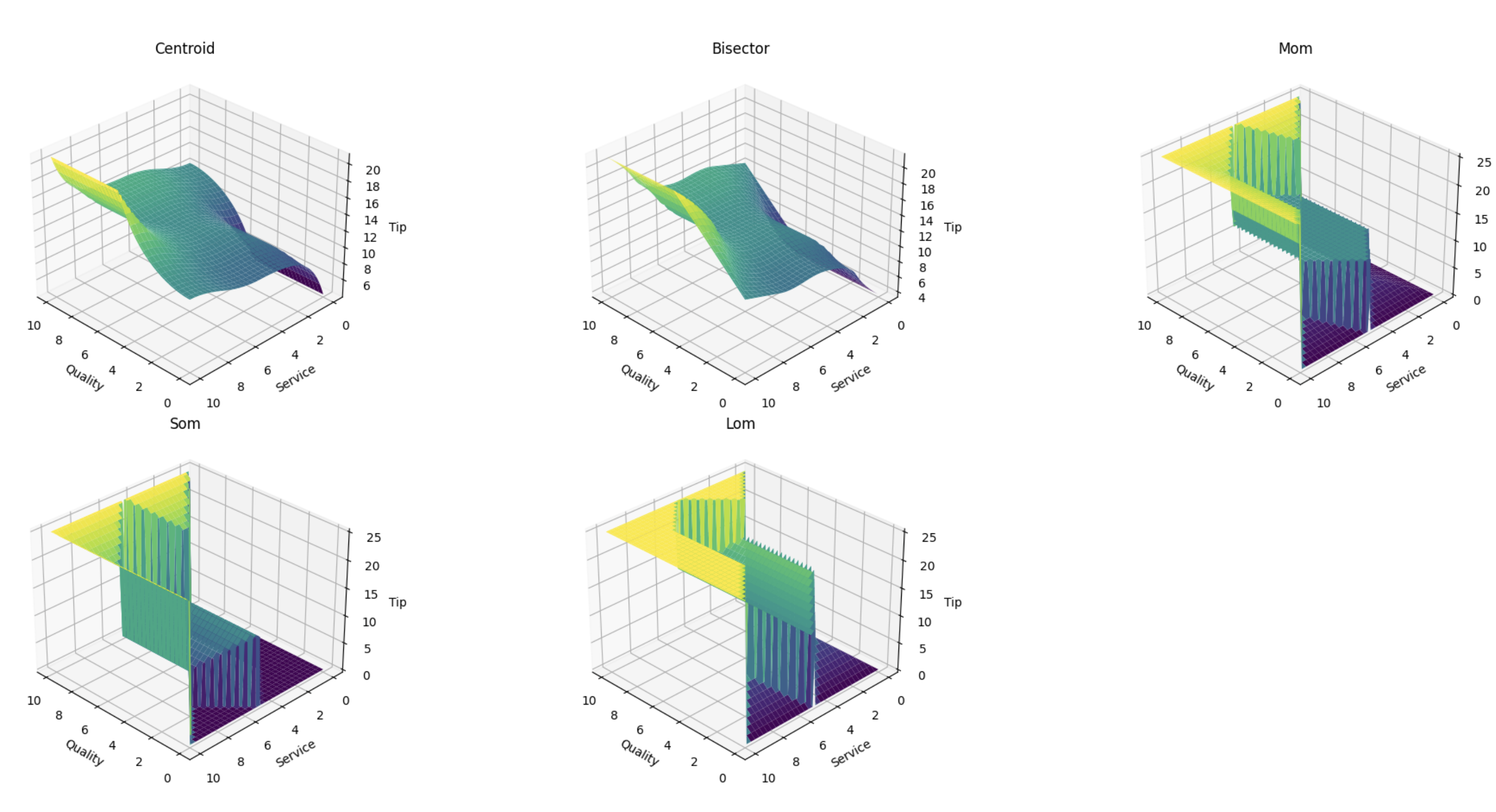}
	\end{center}
	\caption{The relationship between the tip, food quality, and service quality by using different defuzzification methods.}\label{defuzz}
\end{figure}

 The expert overseeing this establishment adheres to specific guidelines when determining the amount of tips to be given, as follows:
\begin{align*}
	R_1:\quad&\text{If (food)  quality is \textit{poor} or service is \textit{poor}, then tip is \textit{low}.}\\
	R_2:\quad&\text{If  service is \textit{average}, then tip is \textit{medium}.}\\
	R_3:\quad&\text{If (food) quality is \textit{good} or service is \textit{good}, then tip is \textit{high}.}
\end{align*}
The ranges for food quality and service quality are both set between $[0, 10]$, whereas the range for the tip is specified as $[0, 25]$. We employ three triangular fuzzy subsets – \textit{poor}, \textit{average}, and \textit{good} – to categorize both food and service qualities. Similarly, for the tip, we utilize three triangular fuzzy subsets: \textit{low}, \textit{medium}, and \textit{high}. 
Then, by employing various defuzzification techniques, the connection between the tip, food quality, and service quality can be demonstrated as depicted in Figure \ref{defuzz}.
Now, we can calculate the values of our fuzzy causal effect formulas using different defuzzification methods as shown in Table \ref{tiptable}. 

\begin{table}[ht]
	\centering
	\begin{tabular}{|c|c|c|c|c|c|}
		\hline
		\diagbox{Causal \\ Formula}{Defuzz \\Method} & Centroid & Bisector & LoM & MoM & SoM\\
		\hline
		$\FATE$ & 3.42 & 3.57 & 9.94 & 9.93 & 9.99 \\
		\hline
		$\NFATE$ & 0.49 & 0.51 & 1.42 & 1.42 & 1.43\\
		\hline
			$\GFATE$ & 2.33 & 2.17 & 3.66 &  3.68 & 3.65\\
		\hline
		$\NGFATE$ & 0.33 & 0.31 & 0.52 &  0.52 & 0.52\\
		\hline
	\end{tabular}
	\caption{The causal effect of food quality on the tip using different fuzzy causal effect formulas and defuzzification methods.}
	\label{tiptable}
\end{table}

In Table \ref{tiptable}, to compute GFATE and NGFATE, it is assumed that  food  quality follows a normal distribution with a mean of 5 and a variance of 2. The unequal NFATE and NGFATE values for each method illustrate the nonlinear relationship between the tip, food quality, and service quality, as depicted in Figure~\ref{defuzz}.
\subsection{Tipping Problem with Probabilistic Rules}
\begin{figure}
	\includegraphics[scale =0.5]{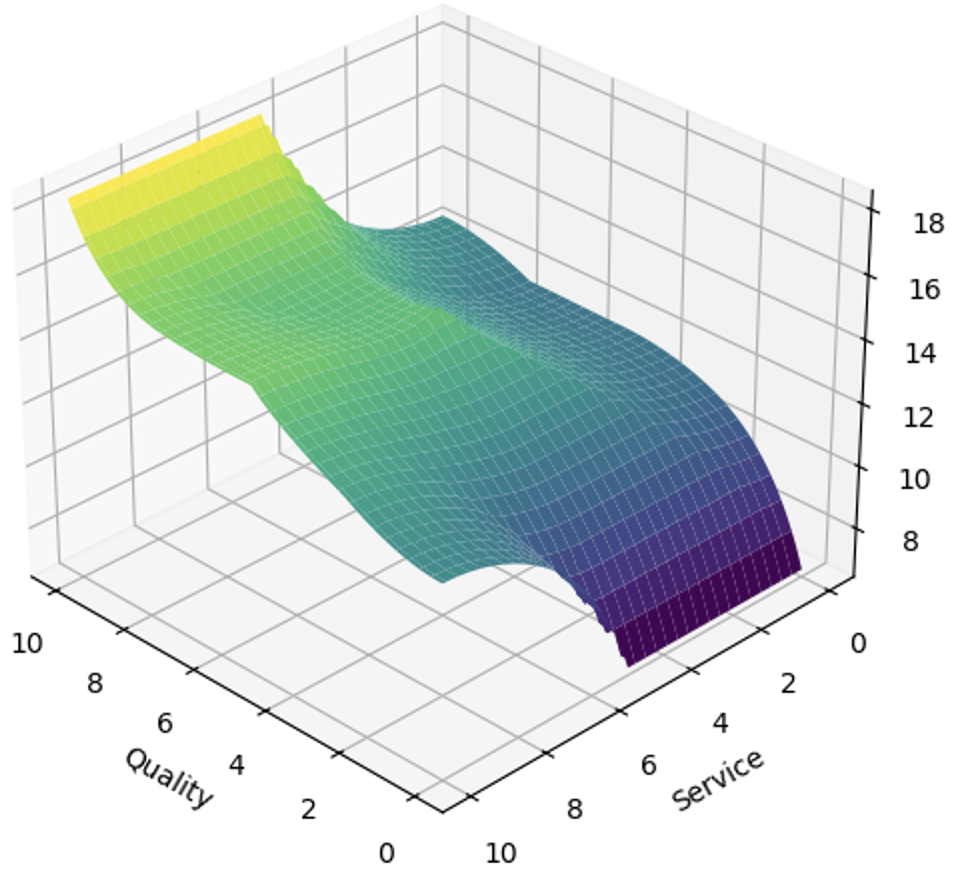}
\caption{Relationship between tipping, food quality, and service using a probabilistic fuzzy system with centroid defuzzification method.}\label{ptp}
\end{figure}

Assume each rule in the tipping scenario occurs with a specific probability. Consider the following setup where each of three antecedents has two potential outcomes:
\begin{align*}
	R_{11}:&\quad \text{If food quality or service is \textit{poor}, then the tip is \textit{low}.} \\
	R_{12}:&\quad \text{If food quality or service is \textit{poor}, then the tip is \textit{medium}.} \\
	R_{21}:&\quad \text{If service is \textit{average}, then the tip is \textit{medium}.} \\
	R_{22}:&\quad \text{If service is \textit{average}, then the tip is \textit{high}.} \\
	R_{31}:&\quad \text{If food quality or service is \textit{good}, then the tip is \textit{high}.} \\
	R_{32}:&\quad \text{If food quality or service is \textit{good}, then the tip is \textit{medium}.}
\end{align*}
Assume \(R_{i1}\) occurs with a probability of 0.7 and \(R_{i2}\) with 0.3. To estimate the causal effect of food quality on tipping, we uphold the assumptions of conditional ignorability and consistency. To compute \(\E[T|q, c]\) for each \(q\) and \(c\), select one rule from each rule pair, estimate the output, and average over all possible rule combinations assuming rule independence. For example, the probability of selecting \(\{R_{11}, R_{22}, R_{32}\}\) is \(0.7 \times 0.3 \times 0.3 = 0.063\). We continue under the presumption that food and service qualities are distributed as described in the standard tipping model.
Furthermore, we utilize only the centroid method to defuzzify the output. Consequently, the following results were calculated as shoen in Table \ref{tab:my_label}:
	\begin{table}
		\centering
			\caption{Estimated causal effects of food quality on tipping using various formulas and a probabilistic fuzzy system with centroid defuzzification.}
		\label{tab:my_label}
		\begin{tabular}{|c|c|c|c|}
			\hline
			FATE   & NFATE & GFATE & NGFATE \\ \hline
			3.03   & 0.43  & 2.21  & 0.32   \\ \hline
		\end{tabular}
	\end{table}
The relationship between variables is demonstrated in Figure \ref{ptp}.
\subsection{Age, Sodium, and Blood Pressure: A Fuzzy Systems Analysis}\label{Age, Sodium, and Blood Pressure: A Fuzzy Systems Analysis}

Returning to the example in Subsection \ref{Sodium Intake and Blood Pressure}, we approach it differently. Instead of directly applying the dataset that captures the interplay among age, sodium intake, proteinuria, and blood pressure, we derive fuzzy rules from this data. Utilizing only these rules, the DAG depicting variable relationships, and the variable value ranges, we create a fuzzy system for blood pressure prediction. We then compute the causal effect of sodium intake and age on blood pressure using our various formulas. 

\begin{figure}
	\begin{center}
		\subfloat[The fuzzy subsets of sodium intake\label{fuzzysubsets}]{\includegraphics[scale =0.35]{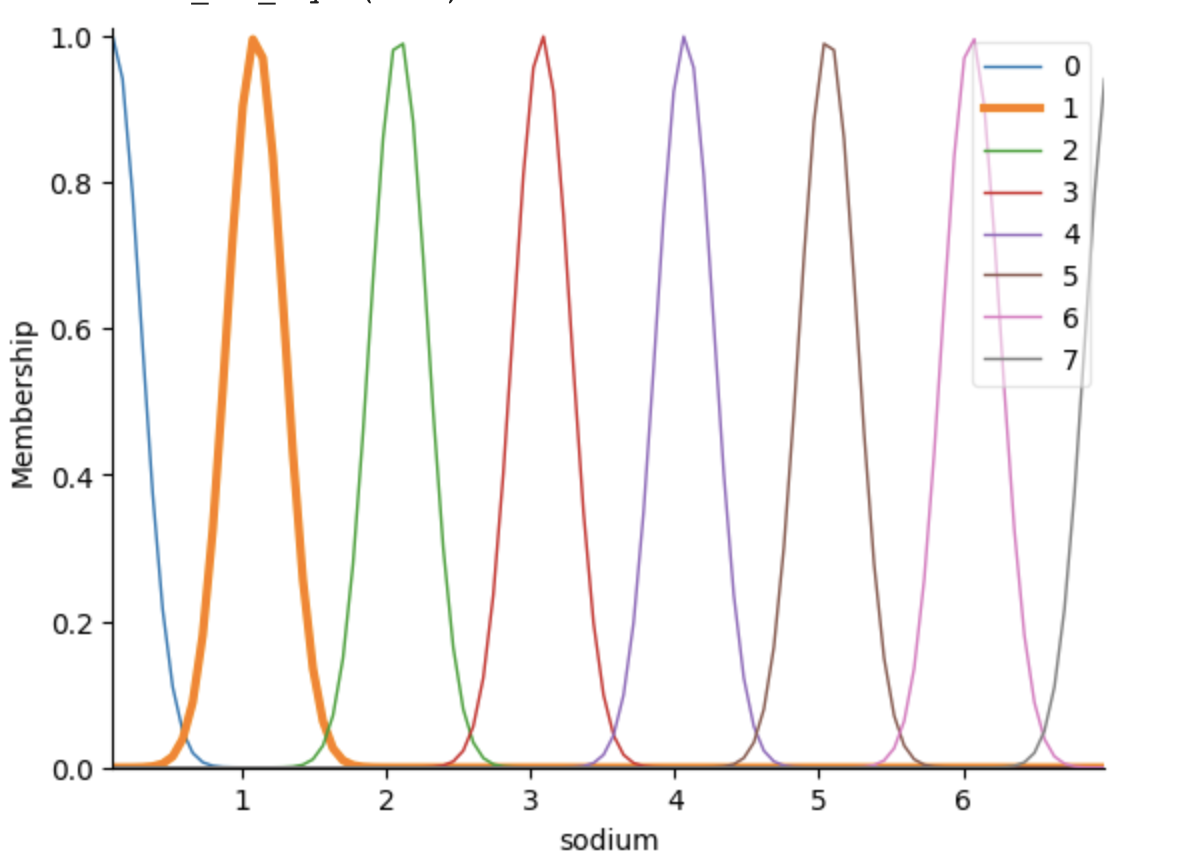}}
		\subfloat[Predicted blood pressure values \label{prediction}]{\includegraphics[scale = 0.35]{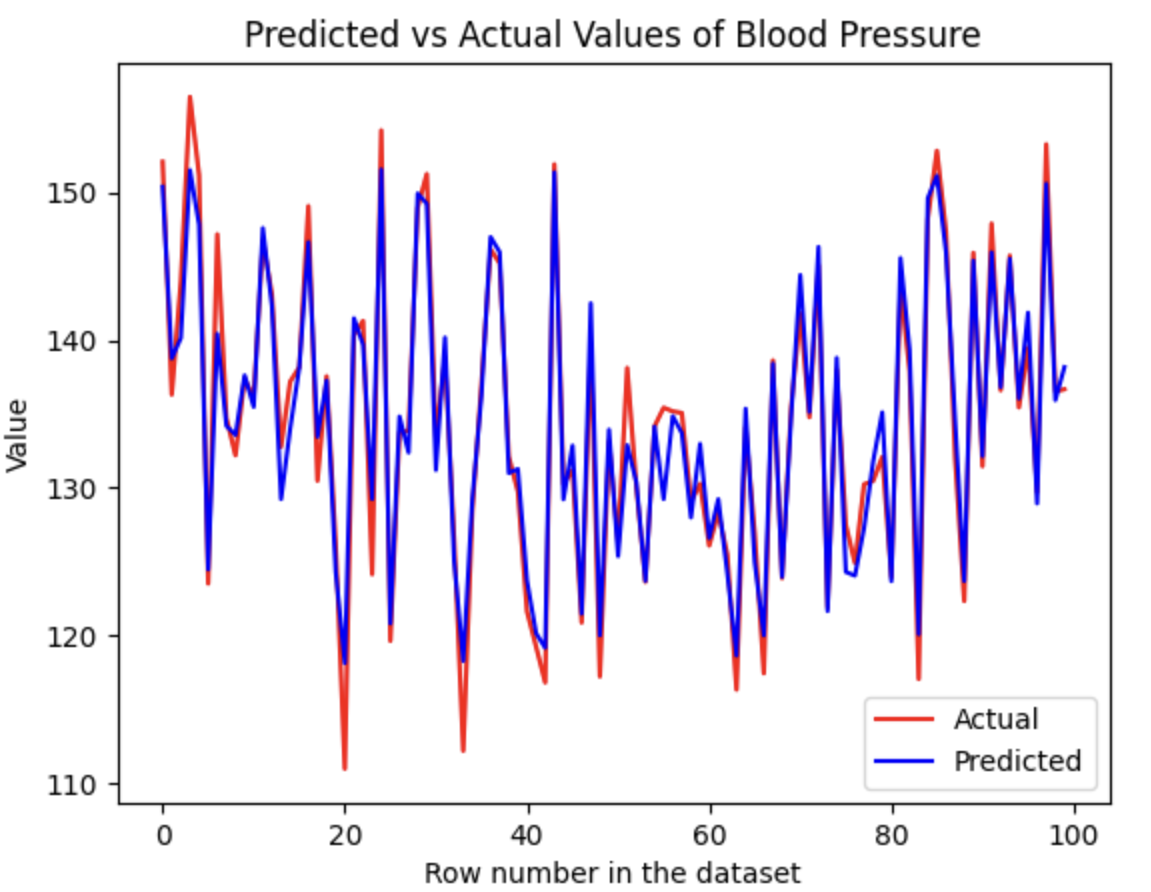}}
		\caption{
			The Gaussian fuzzy subsets of sodium intake are depicted in \ref{fuzzysubsets}. Eight similar fuzzy subsets are also employed for both blood pressure and age. In \ref{prediction}, we present the predicted blood pressure values for the initial 100 entries of the dataset. These predictions are made using a fuzzy control system that incorporates rules derived from the Apriori algorithm. \label{mainexample}}
	\end{center}
\end{figure}

In our python code, we employ a function that accepts a positive integer $n$, representing the number of fuzzy subsets for each variable involved. We utilize Gaussian membership functions in this process. Our approach involves applying these membership functions to the dataset and substituting each data value with the index of the fuzzy subset that assigns the highest membership degree to that value. Subsequently, we use the Apriori algorithm to identify the dominant fuzzy rules within the dataset. We then align these extracted rules with the structure required by the skfuzzy library's fuzzy system. Following this, we predict blood pressure values using the same age and sodium intake data from the dataset. As depicted in Figure \ref{mainexample}, the prediction accuracy with $n=8$ is notably high. The Gaussian nature of the considered  fuzzy sets  facilitates the activation of fuzzy rules. This is due to the Gaussian sets' inherent characteristic of partially overlapping, ensuring that for any given input, there is likely to be some degree of membership in multiple sets. This overlap allows for smoother transitions between rules and a broader range of applicability. In contrast, if triangular fuzzy sets were used, there might be scenarios where some inputs do not activate any rules. This situation can occur because triangular sets have less overlap and are more distinct, potentially leading to gaps where inputs fall outside the defined ranges, resulting in errors or unhandled cases in the fuzzy system. Indeed, while there are methods to determine the most appropriate rule to activate in situations where no rule is initially triggered (\cite{CHEN20131,hajek2013metamathematics, HAONAN2021114504, 9294179, sugeno1993fuzzy,  zimmermann2001fuzzy}), Gaussian fuzzy sets effectively meet our requirements in this experiment. 

The calculated causal effects of age and sodium intake, as determined by various formulas, are presented below:

\begin{table}[ht]
	\centering
	\Small
	\begin{tabular}{|c|c|c|c|c|c|c|c|}
		\hline
		Formula &$\begin{array}{c}\text{True ATE,}\\ \text{NFATE and NGFATE}\end{array}$ & True FATE & ATE & FATE  & NFATE &   GFATE & NGFATE \\
		\hline
		Sodium & 1.05 &4.86 & 1.02 & 4.04 & 1.02 & 2.92 & 1.02\\
		\hline 
		Age & 2 & 50.27 & 2 & 43.08 &  2 & 27.17& 2\\
		\hline 
	\end{tabular}
	\caption{The true values and the estimated values for the ATE, FATE, NFATE, GFATE, and NGFATE of sodium intake and age  on blood pressure.}
	\label{table}
\end{table}
Note that the difference between the true values of FATE in Table~\ref{table} and Table~\ref{your-label-here} is due to the
 variations in the minimum and maximum values of the data generated, as illustrated in Figure \ref{simple}.
\begin{Remark}
In the example presented in Section \ref{Age, Sodium, and Blood Pressure: A Fuzzy Systems Analysis}, the difference between the actual FATEs and the estimated ones arises because we used a fuzzy system without  the exact dataset. Nevertheless, these estimations are notably good. A more coherent approach to employing fuzzy systems here would involve fuzzifying the causal metrics and then interpreting their values as low', medium', `high', or other similar fuzzy attributes.
\end{Remark}
\begin{Remark}
In the example discussed in Section \ref{Age, Sodium, and Blood Pressure: A Fuzzy Systems Analysis}, the complexity of the employed Apriori algorithm is $O(2^pn)$, where $p$ is bounded by $k \times m$. Here, $m$ and $k$ denote the  number of columns and the  number of fuzzy sets employed, respectively. Therfore, the execution time of the code associated with this example could become substantially high if either the number of columns or the number of fuzzy sets is high.
\end{Remark}
\section{Conclusion}

This research has successfully demonstrated the integration of fuzzy logic into causal inference by developing the fuzzy causal metrics, including the Fuzzy Average Treatment Effect (FATE) and the Generalized Fuzzy Average Treatment Effect (GFATE), along with their normalized counterparts as generalizations of classic Average Treatment Effect. 
 Indeed, FATE considers all values of the treatment variable equally important. In contrast, GFATE  takes into account the rarity and frequency of these values.
 
Further, we established the identifiability criteria for the aforementioned metrics to overcome the problem of missing values or counterfactuals in observational studies.

 Furthermore, we proved the stability of both FATE and GFATE under minor perturbations in the data, confirming their practicality for empirical research.
 This study also  provides  robust tools for quantifying causal effects in environments where traditional methods fall short (for instance, in environments where we understand general fuzzy rules between variables rather than precise relationships, such as exact joint distributions or functional relationships). 
 
 Our methodology not only provide a mathematical framework required for causal analysis but also enhance the ability of causal inference techniques to deal with complex and uncertain scenarios especially in fields such as economics and health sciences. The experimental validations emphasize  the reliability of the proposed fuzzy causal inference framework, marking a significant contribution to  statistical analysis. This integration of fuzzy logic and causal inference enhances the  researchers' ability  to obtain more  reliable insights for the task of decision-making across various domains of the sciences.

\section*{Declaration of Generative AI and AI-assisted Technologies in the Writing Process} We used ChatGPT, for grammar checking, spell checking, and partally for enhancing the readability of the manuscript. These tools were used under close human supervision and control. The author(s) have reviewed and edited all AI-generated content and assume full responsibility for the content of the publication.

\bibliographystyle{amsplain}
\bibliography{References}

\providecommand{\bysame}{\leavevmode\hbox to3em{\hrulefill}\thinspace}
\providecommand{\MR}{\relax\ifhmode\unskip\space\fi MR }
\providecommand{\MRhref}[2]{%
  \href{http://www.ams.org/mathscinet-getitem?mr=#1}{#2}
}
\providecommand{\href}[2]{#2}
\begin{thebibliography}{10}

\bibitem{al2014automatic}
Mohammed Al-Shammaa and Maysam~F Abbod, \emph{Automatic generation of fuzzy
  classification rules from data}, Proc. of the 2014 International Conference
  on Neural Networks-Fuzzy Systems (NN-FS'14), Venice, 2014, pp.~74--80.

\bibitem{CHEN20131}
Shyi-Ming Chen and Cheng-Yi Wang, \emph{Fuzzy decision making systems based on
  interval type-2 fuzzy sets}, Information Sciences \textbf{242} (2013), 1--21.

\bibitem{driankov2013introduction}
Dimiter Driankov, Hans Hellendoorn, and Michael Reinfrank, \emph{An
  introduction to fuzzy control}, Springer Science \& Business Media, 2013.

\bibitem{ductu2017fast}
Liviu-Cristian Du{\c{t}}u, Gilles Mauris, and Philippe Bolon, \emph{A fast and
  accurate rule-base generation method for mamdani fuzzy systems}, IEEE
  Transactions on Fuzzy Systems \textbf{26} (2017), no.~2, 715--733.

\bibitem{faghihi2020association}
Usef Faghihi, Serge Robert, Pierre Poirier, and Youssef Barkaoui, \emph{From
  association to reasoning, an alternative to pearls’ causal reasoning}, The
  Thirty-Third International Flairs Conference, 2020.

\bibitem{fakhrahmad2007constructing}
Seyed~Mostafa Fakhrahmad, A~Zare, and M~Zolghadri Jahromi, \emph{Constructing
  accurate fuzzy rule-based classification systems using apriori principles and
  rule-weighting}, Intelligent Data Engineering and Automated Learning-IDEAL
  2007: 8th International Conference, Birmingham, UK, December 16-19, 2007.
  Proceedings 8, Springer, 2007, pp.~547--556.

\bibitem{hajek2013metamathematics}
Petr H{\'a}jek, \emph{Metamathematics of fuzzy logic}, vol.~4, Springer Science
  \& Business Media, 2013.

\bibitem{HAONAN2021114504}
Si~Haonan, Shao Xingling, and Zhang Wendong, \emph{Fuzzy rule-based neural
  appointed-time control for uncertain nonlinear systems with aperiodic
  samplings}, Expert Systems with Applications \textbf{170} (2021), 114504.

\bibitem{kim1999efficient}
Myung~Won Kim, Joong~Geun Lee, and Changwoo Min, \emph{Efficient fuzzy rule
  generation based on fuzzy decision tree for data mining}, FUZZ-IEEE'99. 1999
  IEEE International Fuzzy Systems. Conference Proceedings (Cat. No.
  99CH36315), vol.~3, IEEE, 1999, pp.~1223--1228.

\bibitem{klir1995fuzzy}
George Klir and Bo~Yuan, \emph{Fuzzy sets and fuzzy logic}, vol.~4, Prentice
  hall New Jersey, 1995.

\bibitem{klir1996fuzzy}
George~J Klir and Bo~Yuan, \emph{Fuzzy sets, fuzzy logic, and fuzzy systems:
  selected papers by lotfi a zadeh}, vol.~6, World Scientific, 1996.

\bibitem{kunitomo2022causal}
Lucie Kunitomo-Jacquin, Aurore Lomet, and Jean-Philippe Poli, \emph{Causal
  discovery for fuzzy rule learning}, 2022 IEEE International Conference on
  Fuzzy Systems (FUZZ-IEEE), IEEE, 2022, pp.~1--8.

\bibitem{9294179}
Le~Hung Lan, Phi Van~Lam, and Nguyen Van~Hai, \emph{An approach to the analysis
  and design of fuzzy control system}, 2020 3rd International Conference on
  Robotics, Control and Automation Engineering (RCAE), 2020, pp.~36--40.

\bibitem{luque2019educational}
Miguel~Angel Luque-Fernandez, Michael Schomaker, Daniel Redondo-Sanchez, Maria
  Jose Sanchez~Perez, Anand Vaidya, and Mireille~E Schnitzer, \emph{Educational
  note: Paradoxical collider effect in the analysis of non-communicable disease
  epidemiological data: a reproducible illustration and web application},
  International journal of epidemiology \textbf{48} (2019), no.~2, 640--653.

\bibitem{miao2000causal}
Yuan Miao and Zhi-Qiang Liu, \emph{On causal inference in fuzzy cognitive
  maps}, IEEE transactions on Fuzzy Systems \textbf{8} (2000), no.~1, 107--119.

\bibitem{nerbass2018sodium}
Fabiana~B Nerbass, Viviane Calice-Silva, and Roberto Pecoits-Filho,
  \emph{Sodium intake and blood pressure in patients with chronic kidney
  disease: a salty relationship}, Blood purification \textbf{45} (2018),
  no.~1-3, 166--172.

\bibitem{ross2009fuzzy}
Timothy~J Ross, \emph{Fuzzy logic with engineering applications}, John Wiley \&
  Sons, 2009.

\bibitem{saki2022fundamental}
Amir Saki and Usef Faghihi, \emph{A fundamental probabilistic fuzzy logic
  framework suitable for causal reasoning}, arXiv preprint arXiv:2205.15016
  (2022).

\bibitem{sakifaghihi}
\bysame, \emph{Exploring simpson’s paradox through the lensof fuzzy causal
  inference}, IEA/AIE, 2024.

\bibitem{schneider2016case}
Carsten~Q Schneider and Ingo Rohlfing, \emph{Case studies nested in fuzzy-set
  qca on sufficiency: Formalizing case selection and causal inference},
  Sociological Methods \& Research \textbf{45} (2016), no.~3, 526--568.

\bibitem{sugeno1993fuzzy}
Michio Sugeno and Takahiro Yasukawa, \emph{A fuzzy-logic-based approach to
  qualitative modeling}, IEEE Transactions on fuzzy systems \textbf{1} (1993),
  no.~1, 7.

\bibitem{wang1992generating}
L-X Wang and Jerry~M Mendel, \emph{Generating fuzzy rules by learning from
  examples}, IEEE Transactions on systems, man, and cybernetics \textbf{22}
  (1992), no.~6, 1414--1427.

\bibitem{wang1994adaptive}
Li-Xin Wang, \emph{Adaptive fuzzy systems and control: design and stability
  analysis}, Prentice-Hall, Inc., 1994.

\bibitem{wang2014fuzzy}
Yingxu Wang, \emph{Fuzzy causal inferences based on fuzzy semantics of fuzzy
  concepts in cognitive computing}, WSEAS Transactions on Computers \textbf{13}
  (2014), 430--441.

\bibitem{weir2012urinary}
Matthew~R Weir, Raymond~R Townsend, Jeffrey~C Fink, Valerie Teal, Stephen~M
  Sozio, Cheryl~A Anderson, Lawrence~J Appel, Sharon Turban, Jing Chen, Jiang
  He, et~al., \emph{Urinary sodium is a potent correlate of proteinuria:
  lessons from the chronic renal insufficiency cohort study}, American journal
  of nephrology \textbf{36} (2012), no.~5, 397--404.

\bibitem{zadeh1965fuzzy}
Lotfi~A Zadeh, \emph{Fuzzy sets}, Information and control \textbf{8} (1965),
  no.~3, 338--353.

\bibitem{zadeh1974fuzzy}
\bysame, \emph{Fuzzy logic and its application to approximate reasoning.}, IFIP
  congress, vol. 591, 1974.

\bibitem{zhou2006fuzzy}
Sanming Zhou, Zhi-Qiang Liu, and Jian~Ying Zhang, \emph{Fuzzy causal networks:
  General model, inference, and convergence}, IEEE Transactions on Fuzzy
  Systems \textbf{14} (2006), no.~3, 412--420.

\bibitem{zimmermann2001fuzzy}
H-J Zimmermann, \emph{Fuzzy data analysis}, Fuzzy Set Theory—and Its
  Applications, Springer, 2001, pp.~277--328.

\end{thebibliography}
\end{document}